# Deep Learning for Automatic Facial Detection and Recognition in Japanese Macaques: Illuminating Social Networks

Julien Paulet[1,2], Axel Molina[3], Benjamin Beltzung[4], Takafumi Suzumura[2], Shinya Yamamoto[2,5], Cédric Sueur[4,6,7]

1   Université Jean Monnet, Saint-Etienne, France
2   Wildlife Research Center, Kyoto University, Kyoto, Japan
3   Ecole Normale Supérieure, Université PCL, Paris, France
4   Université de Strasbourg, IPHC UMR7178, CNRS, Strasbourg, France
5   Kyoto University Institute for Advanced Study, Kyoto, Japan
6   ANTHROPO-LAB, ETHICS EA 7446, Université Catholique de Lille, Lille, France
7   Institut Universitaire de France, Paris, France

**Abstract:**

Individual identification plays a pivotal role in ecology and ethology, notably as a tool for complex social structures understanding. However, traditional identification methods often involve invasive physical tags and can prove both disruptive for animals and time-intensive for researchers. In recent years, the integration of deep learning in research offered new methodological perspectives through automatization of complex tasks. Harnessing object detection and recognition technologies is increasingly used by researchers to achieve identification on video footage. This study represents a preliminary exploration into the development of a non-invasive tool for face detection and individual identification of Japanese macaques (*Macaca fuscata*) through deep learning. The ultimate goal of this research is, using identifications done on the dataset, to automatically generate a social network representation of the studied population. The current main results are promising: (i) the creation of a Japanese macaques' face detector (Faster-RCNN model), reaching a 82.2% accuracy and (ii) the creation of an individual recognizer for Kōjima island macaques population (YOLOv8n model), reaching a 83% accuracy. We also created a Kōjima population social network by traditional methods, based on co-occurrences on videos. Thus, we provide a benchmark against which the automatically generated network will be assessed for reliability. These preliminary results are a testament to the potential of this innovative approach to provide the scientific community with a tool for tracking individuals and social network studies in Japanese macaques.

**Introduction**

In the fields of ethology and ecology, individual identification plays a crucial role, enabling a wide range of research domains. Long-term observations spanning days, months, or even years, rely on the ability to track and differentiate individuals thus enabling to follow-up some variables while ensuring the stability of others (Clutton-Brock & Sheldon, 2010). Individual identification is pivotal in various fields and scales, including the study of social group dynamics, heredity of traits through lineage, phenotypic changes in one in response to life history, or behavioral transmission in a population (Clutton-Brock & Sheldon, 2010; Moss, 2001; Wich et al., 2004). Moreover, beyond its role in fundamental ethological and ecological research, individual identification holds significant importance in applied science and conservation efforts. This allows to minimize losses during problematic species regulation by targeting specific individuals. It can also be used to cater for the specific needs of protected species (Sinha et al., 2018). Our capacity to accurately differentiate and track individuals then appears as heavily impacting.

Various methods of identification have been employed since we try to identify and track individuals. We can cite banding, collaring or tagging, but also the attachment of GPS trackers or other kinds of emitter directly on the body (e.g., in marine mammals: Walker et al., 2011, in bears: Bethke et al. 1996, in primates : Honess & Macdonald, 2003). However, most of these methods can be invasive and potentially detrimental to animals. They can lead to injuries during human intervention, induce stress, inhibit movement, and visually differentiate individuals from their conspecifics, thereby altering behavior and introducing biases into the studies. These invasive methods also raise ethical concerns and compromise animal welfare (Hermon & Sharma, 2021; Zemanova, 2020)

Thus, visual biometric identification emerges as a highly preferable method for individual identification in ecology. It requires no costly or challenging-to-use instruments and minimizes the need to disturb the animals. Additionally, visual identification is applicable to a wide range of species, as long as there are at least subtle inter-individual phenotypic variations. By adopting biometric individual identification, researchers can save valuable resources, enhance the reliability of ethological observations and data collection, and ensure a more ethical approach (Zemanova, 2020). From those perspectives, it appears to be the most advantageous approach.

While ecological and behavioral data can sometimes be collected directly on the field, studies based on individual recognition may generally require video recording because of complex and fast behavioral sequences. Video recordings offer the advantage of delayed analysis and extensive data collection. This method facilitates longitudinal studies and enables precise tracking of individuals' life histories, social dynamics within a group, or species abundance and distribution in a given area (Caravaggi et al. 2017). Consequently, large databases have been created for some conservation and ethological projects, providing highly valuable resources for scientific research (e.g., in mandrills, *Mandrillus sphinx*: Tieo et al., 2023)

However, the use of those databases is challenged by the very large volume of data to be processed before analysis. Notably, individual recognition typically demands a significant investment of time, effort, training, and expertise. While social animals generally possess the ability to visually recognize and distinguish others in their own species, certain visual traits may prove challenging for an untrained human observer to identify accurately (Kühl & Burghardt, 2013). Human observer identification training induces consequent time consumption in a research project, even more when multiple researchers are involved in the data collection and analysis. These challenges limit the potential of long-term recorded data and hinder research based on individual identification.

To address these challenges, automated methods for individual identification have emerged in recent years, primarily driven by advancements in computer vision and algorithms. This has led to a methodological paradigm shift in ecology through the application of artificial intelligence (AI) trained using machine learning and deep learning techniques. Deep learning empowers AI systems to detect patterns and complex interactions often difficult for human analysis to access. Unlike traditional programming approaches, deep learning automatically creates models with multiple layers of processing by learning from examples, enabling the system to extract meaningful information through deep abstract manipulation of complex input (Valletta et al., 2017). In ethology, most studies concern uncovering the cryptic and complex mechanisms underlying observable behaviors. This is why deep learning may slowly become a new natural framework of approach in ethological problems, gradually unveiling new perspectives and insights in this discipline (see Valletta et al., 2017 and Weinstein, 2018 for examples of deep learning applications regarding complex morphological data analysis, social network analysis or individual identification).

Particularly in the realm of image analysis and object recognition, AI has become a highly used tool in various human activities (Zhang & Lu, 2021). In the field of biology, image analysis and object detection are among the first applications of deep learning, for species identification based on phenotypic biometrics. Over the past decade, there has been a surge in projects focused on automated individual recognition in animals (e.g. in right whales, *Eubalaena glacialis*: Bogucki et al., 2019; in farm pigs, *Sus domesticus*: Hansen et al., 2018; in giant pandas, *Ailuropoda melanoleuca*: Hou et al., 2020; in Amur tigers, *Panthera tigris altaica*: Shi et al., 2020; in the small birds *Philetairus socius*, *Parus major* and *Taeniopygia guttata*: Ferreira et al., 2020); in giraffes, *Giraffa camelopardalis*: Miele et al., 2021; in Asian elephants, *Elephas maximus*: de Silva et al., 2022) having an impact on species conservation and biodiversity monitoring. Going deeper in identification precision researchers explored the possibility of individual identification by face analysis in some selected species, just as applications of face detection have proliferated and diversified for humans in recent years. Focusing on the face offers a significant advantage due to the distinctive inter-individual variations exhibited by facial traits, particularly in primates, where these traits are generally stable and recognizable across various life events and time, akin to humans (Schofield et al., 2023).

Several projects have already been conducted with different primate clades (in lemurs: Crouse et al., 2017; in mandrills, *Mandrillus sphinx*: Charpentier et al., 2020; in western lowland gorillas, *Gorilla gorilla gorilla*: Brookes et al., 2022, and in 41 different species: Guo et al., 2020). Notably, Schofield et al. (2023) developed a fully automated pipeline for face detection, tracking, and recognition of 23 chimpanzees in the long-term field site of Bossou in Guinea. This project uses face recognition directly from video data. Unlike traditional AI approaches that rely on single frames, this recognition method tracks faces across multiple frames to make predictions, enhancing the reliability and certainty of individual identification. Additionally, their AI pipeline has been employed to generate co-occurrence matrices of individuals based on hours of video footage, providing insights into the social network of group members and its changes over time. Claire L. Witham also employed automatic face recognition to generate a social network in rhesus macaques in 2018, although her model relied on manual training rather than deep learning techniques.

In the present study, drawing from Schofield et al. work, we aim to investigate whether similar deep learning-based techniques for face detection, recognition, and subsequent automatic construction of a social network can be applied on Japanese macaques (*Macaca fuscata*). Japanese macaques are an endemic monkey species inhabiting various ecosystems across Japan. They typically form troops consisting of approximately 30 individuals, with around two-thirds being females and one-third males (Sugiyama, 1976). Troops of macaques exhibit a well-defined and strict social hierarchy. Females generally inherit their rank, while males engage in competition to establish hierarchy (N. Koyama, 1967; N. F. Koyama, 2003). These troops follow a matrilineal society, where females remain in their natal group throughout their lives (Fedigan & Asquith, 1991), while males disperse at sexual maturity, typically around four years old (Nakamichi, 1989; Sugiyama, 1976) , and may undergo emigration from their troop during mating seasons several times in their life. They practice polygamous mating, males and females typically having multiple partners during the mating season (Solthis, 1999). Social cohesion within troops is primarily maintained through grooming behavior, which is most prevalent among females and juveniles (Baxter & Fedigan, 1979; N. Koyama, 1967; Nakamichi & Shizawa, 2003).

Japanese macaques hold a significant ethological interest as they exhibit a rich and subtle socio-cognitive life. They were one of the first species to provide insights on the concept of animal culture, demonstrating the emergence and transmission of behaviors such as sweet potato washing (Kawai, 1965). They are still often studied for their social behaviors, communication skills, community management through hierarchy, cultural behaviors development and transmission (e.g., Huffman et al., 2010; Rebout et al., 2020; Shimada & Sueur, 2018). Since this species is often studied directly in free-ranging populations, with a strong interest in follow-up of social interactions, reliable individual identification is generally mandatory, but also complex. Therefore, investing in development of automatic identification tools can provide substantial support for future Japanese macaque research. Recent studies by Otani and Ogawa (2021) on 11 Yakushima Island macaques and Ueno et al. (2021) on 51 Katsuyama macaques have already explored some potential of individual identification in Japanese macaques using different learning methods for their models. Here we try to go further by building a freely available automatic system with social network visualization for the Kōjima macaque population, which is one of the most studied in Japan.

Building upon those previous works, our goal is ultimately to create an automatic pipeline capable of detecting Japanese macaques' faces, tracking them across video frames, recognizing individuals from a known set of macaques, and finally constructing the social structure of the population. The work we present here is a methodological exploration of the following three main objectives: (i) the development of a face detector able to automatically localize Japanese macaque faces on video frames; (ii) the development of an individual recognizer for Kōjima macaques, able to identify the detected faces; (iii) an analysis of the Kōjima macaques social network, which will serve to evaluate our future pipeline's performance to automatically generate social network based on recognized macaques' identities. Furthermore, this research project presents an opportunity to contribute to the first face labeled dataset of videos featuring the main population of Japanese macaques on Kōjima Island. Ultimately, our project aims to provide a reliable tool for tracking individuals, enabling a deeper understanding of how macaques form social networks and gaining insights into processes such as the social transmission of cultural behaviors.

**Material and methods**

A graphical representation of our methodological approach is presented in Figure 1, summarizing the realized work and the perspectives still to explore.

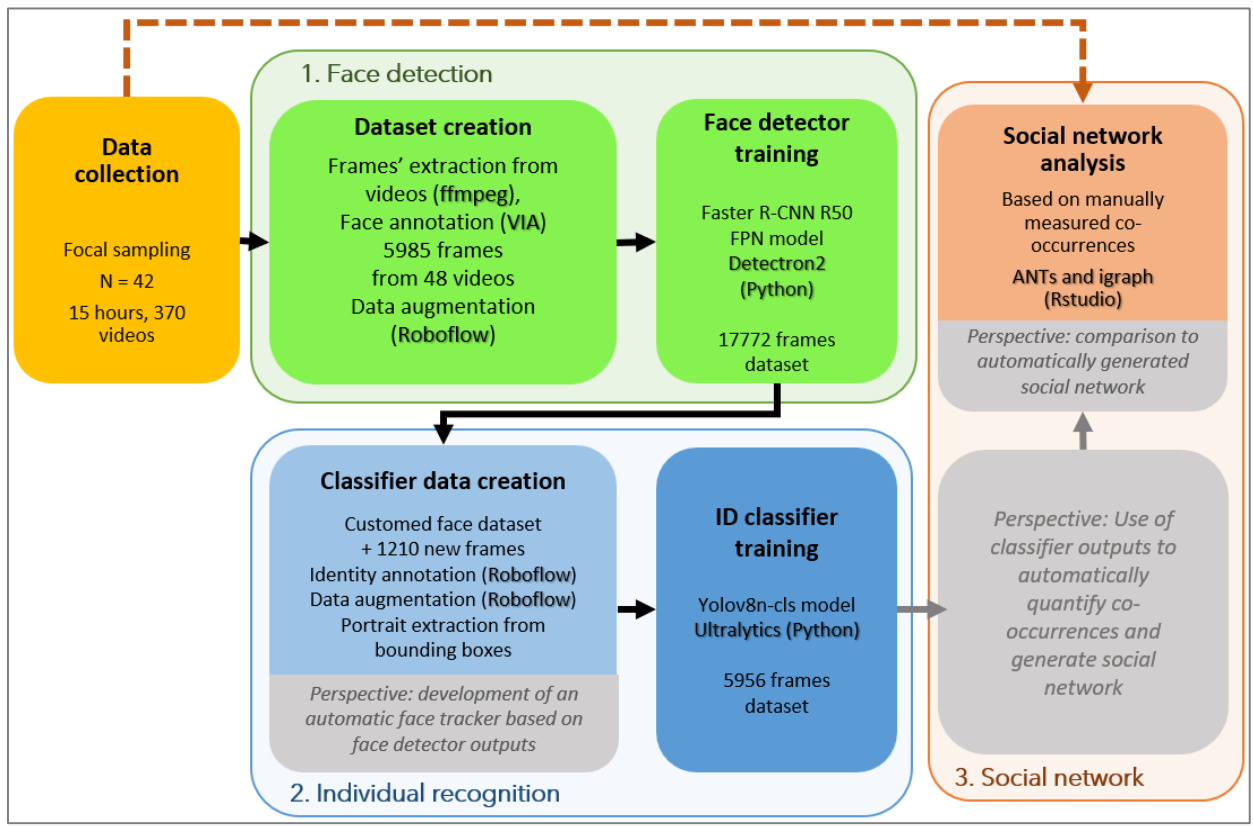

Figure 1. Methodological graphic summary regarding the project three main objectives. Bold text within colored boxes relates to software, frameworks and packages we used.

*Study group and localization*

The study took place on Kōjima, a small island spanning approximately 30 hectares situated in the Sea of Hyūga, about 300 meters off the Nichinan Coast, near the city of Kushima in Miyazaki Prefecture. The island is mainly boarded by rocky coasts and has one sandy beach called Odomari (Iwamoto, 1974) sinking for 100 meters into the island (**Figure 2.A**). Kōjima Island serves as a field study site affiliated with the Wildlife Research Center at Kyoto University and is hosting a community of approximately 100 Japanese macaques.

These macaques are divided into three distinct populations: the 'main group,' the 'maki group,' and the 'hitorizaru' which are free-ranging males. Each individual on the island is named and documented in a maternal pedigree spanning over 60 years. To build our dataset, we chose to focus specifically on the 'main group,' which is regularly the subject of extensive research. While the macaques on the island live and behave freely, the members of the 'main group' come daily around the beach, essentially in the morning (**Figure 2.B**). Researchers provide them with soybeans as supplemental feeding, which occurs one to three days per week, depending on weather conditions impacting accessibility to the island. The macaques are periodically weighed by baiting them onto scales with food (**Figure 2.C**). A few tourists visit the beach regularly to observe the monkeys. Consequently, the Kōjima macaques have become habituated to human presence and appear to exhibit minimal reactions.

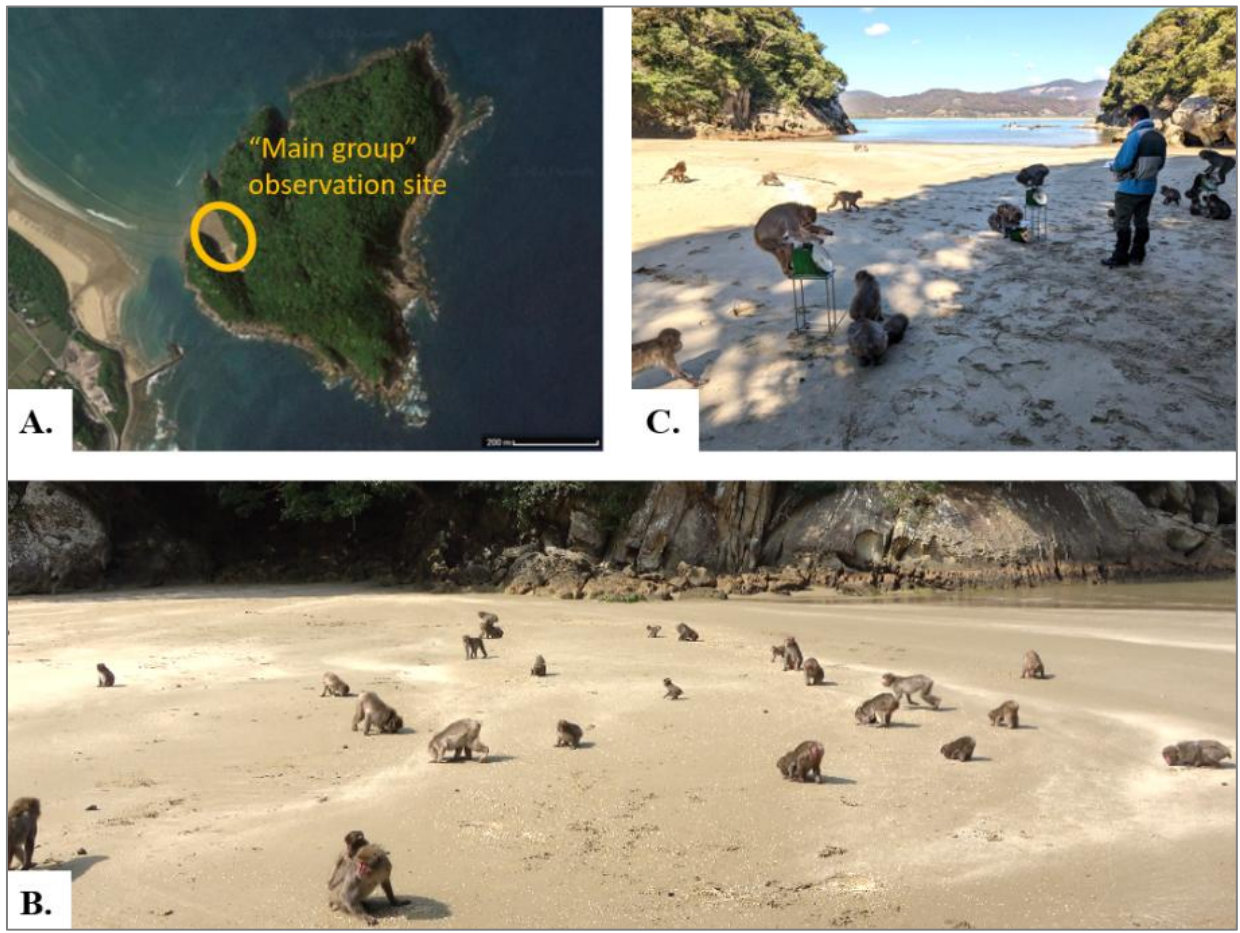

**Figure 2. Contextual pictures of Kōjima and its 'main group' population.** (A) Satellite shot of Kōjima (Source: Google map), (B) 'Main group' macaques feeding on the beach, (C) Baited macaques on scales.

During the two-month observation period from February to March 2023, the 'main group' comprised 42 individuals, including 26 females, 11 males, and 5 yet-to-be-sexed infants (**Supplementary Material**). Occasionally, three free-ranging males would join the group. It is important to note that the observation period coincided with the mating season. Throughout most of the observation sessions, the current alpha male of the 'main group,' Shika, was absent, likely engaging in mating with females from the other group. Muku, the oldest female in the group, passed away during the middle of the observation period.

*Data collection*

During the two-month period from February to March 2023, the Kōjima 'main group' of macaques was subjected to daily filming sessions, contingent upon favorable weather conditions that allowed for boat transportation to the island. The camcorder used was a Sony Handycam 4K AX43A. Upon our arrival, the macaques were promptly provided with food, assuring most of the population to be attracted to the beach. Following the feeding session, the macaques would generally settle near their relatives and partners on the beach, engaging in regular bouts of grooming. To facilitate data collection and maximize reliability, most recording was achieved during resting periods. Each group member was systematically identified and individually filmed for an approximately two and a half minutes focal sampling, following a pre-established randomized list that varied for each session. The primary objective during the video recording was to capture an extended duration of the focused individual's front face and to diversify the facial viewpoints as much as possible.

In instances where the focused individual was in 'social proximity' with other group members (within a one-meter radius from the focused individual), the camera zoom was adjusted to encompass the surrounding individuals, ensuring their visibility on the recorded footage. The zoom level and filming method were calibrated to include only those individuals in proximity, while ensuring that the faces of all individuals were the most clearly discernible in the footage.

Throughout the two-month observation period, detailed records were maintained to monitor the duration and quality of footage obtained for each individual. During last weeks, special attention was given to scaling the quantity and quality of recordings for each group member, ensuring a balanced dataset. Subsequently, the recorded footage was carefully reviewed with Takafumi Suzumura, working on the site since tens of years, to ensure accurate identification of individuals featured in the recordings. The total amount of data is composed of 370 videos for around 15 hours recorded.

*Dataset Annotation for Face Detector*

To develop the macaques' face detector, a dataset of annotated frames was created for training a deep learning model. A total of 5985 frames were extracted from 48 videos, with a sampling rate of one frame per second, using the open-source software ffmpeg (Tomar, 2006). The selection of videos aimed to ensure a balanced representation of each individual and a diverse range of backgrounds, camera angles, and behaviors. The annotations were performed using the web-based tool VGG Image Annotator. The annotation process involved tracing bounding boxes labeled as 'macaque' around the faces of each macaque. These annotations met specific criteria for face detection training, such as faces oriented between frontal and profile views, with a maximum obstruction of 30% and without blurriness or small faces at far distance in the background.

Out of the 5985 frames in the dataset, 3011 frames contained at least one good quality macaque face, while 2974 frames had no annotations. The annotated frames comprised a total of 3555 face bounding boxes. To this dataset were added 642 frames extracted at a rate of 1 frame per second from the 'Japanese Macaques Look Almost Human' YouTube video by Animalogic, generating a 6622 frames dataset. These frames were annotated following the same methodology as the initial dataset frames. This augmentation of the dataset introduces multiple distinct faces not found on Kōjima Island, enhancing the dataset's diversity and maximizing the potential for robust generalization in Japanese macaque face detection. Generalization refers to a model's ability to perform well on new, unseen data that was not part of the training set.

The information about bounding boxes' position on frames are organized at COCO format and exported via a JavaScript Object Notation (json) file to be readable for the detector training.

*Data Augmentation*

Subsequently, the dataset underwent a data augmentation process. Data augmentation is the generation of several new frames by applying modifications on frames already in the dataset. Data augmentation involves generating numerous new frames through the application of modifications to existing frames within the dataset. A wide spectrum of modifications is possible (e.g., image flipping, colorimetric adjustments, grayscale conversion), frequently resulting in an expansion of the initial dataset size by more than twofold. The primary rationale behind the application of data augmentation in this context is to introduce variations and amplify dataset diversity, thereby fostering enhanced generalization during model training and mitigating the risk of overfitting (e.i. when a model becomes excessively specialized to the training data and fails to generalize effectively to novel examples). By incorporating data augmentation techniques, we introduce variations that help the model learn the underlying patterns present in the data, rather than memorizing specific instances.

The execution of data augmentation was carried out using the computer vision software Roboflow (Dwyer et al., 2022). The following modifications have been selected: 90° rotation clockwise and counter-clockwise, 180° rotation, rotation between -45° and +45°, shear ±45° horizontal and ±45° vertical, apply grayscale to 25%, saturation between -60% and +60%, brightness between -25% and +25%, exposure between -10% and +10%, blur up to 1px. Three new images were generated per frame in the training set using randomly selected modifications from this list. Leveraging data augmentation, the final dataset size for training the face detection model reached 17772 frames.

*Face Detection*

For the task of face detection, a Convolutional Neural Network (CNN) model was trained using deep learning. The foundation of this model stems from the Faster R-CNN architecture, featuring a ResNet-50 backbone, and it is constructed atop a Feature Pyramid Network (FPN) architecture. Prior to training on our specific task, the model underwent pre-training through three iterations on the COCO dataset, primarily designed for object detection tasks. Following these pre-training phases, the model underwent fine-tuning using our specialized macaque face detection dataset.

Our dataset, comprising 17772 frames, was partitioned into two distinct subsets. The training set of 16266 frames, constituting around 92% of the dataset, was dedicated to training the model, while 1506 frames, equivalent to 8% of the dataset, were allocated to form the validation set. This validation subset served as a benchmark to evaluate the model's performance all along the training.

The implementation of the model was realized through the utilization of Detectron 2 (Wu et al. 2019), an open-source object detection framework developed atop PyTorch. This framework facilitated the seamless orchestration of the complex training process. The coding environment employed for this undertaking was Google Colab, a Jupyter notebooks service, which provided a robust and flexible platform for the model development. Training the model necessitated substantial computational resources. The process was conducted on Google Colab's Backend Google Compute Engine Python 3 virtual machine service, benefiting from the computational prowess of the A100 GPU. Hyperparameters were tuned to achieve optimal results within our computational capacities. Those are diverse settings that influence the deep learning algorithm during training. The training process have been made with a batch size of 8 frames per batch, denoting the number of files processed concurrently before the model weights are updated. The learning rate was set at 0.001, influencing the gradient descent algorithm's pace during weight updates. The training persisted for 6000 batches.

*Dataset creation for Individual Classifier*

For the individual classifier training, we used the initial 5985 frames dataset randomly reduced by half and supplemented with 1210 new frames manually extracted from a wide variety of videos from all the videos repertory. As specific individual classification requires a greater diversity of examples for each individual, this reduction/supplementation was made to decrease similar frames and increase new ones in the dataset to ensure a better performance. The selection of new frames was done by selecting specific frames in videos to enrich the initial dataset with new point of views, facial expression and backgrounds for each individual. Those new macaques' faces were then annotated following the same method as for face detector dataset, directly on Roboflow to ease its integration in the initial dataset. All the bounding boxes have been manually labeled with the corresponding macaque's name as a class. All the frames from the initial dataset with no bounding box were deleted, as they only were necessary for face detection training. The dataset for individual classifier, before data augmentation, contains 2485 frames with 42 classes, corresponding to the 42 individuals' names.

A program was made using the image processing library 'OpenCV-Python' to extract normalized 240x240 pixels images from the faces bounding boxes. This collection of portraits was then data augmented using roboflow. The following modifications have been selected: 90° rotation clockwise and counter-clockwise, 180° rotation, rotation between -15° and +15°, shear ±15° horizontal and ±15° vertical, apply grayscale to 5%, saturation between -25% and +25%, brightness between -6% and +6%, exposure between -6% and +6%, blur up to 1px. Three new images were generated per frame in the training set using randomly selected modifications from this list. The resulting dataset contained 5956 images.

*Individuals Classifier*

For the task of individual recognition, a Convolutional Neural Network (CNN) model was trained using deep learning. This model follows a 'You Only Look Once' (YOLO) architecture with the v8 Nano extension (e.i. the fastest model version for YOLO architecture) especially pretrained for classification with ImageNet.

Our dataset, comprising 5956 frames, was partitioned into three distinct subsets. The training set contained 5205 frames, constituting around 87% of the dataset, the validation set contained 494 frames, equivalent to 8% of the dataset, and the test set contained 248 frames, constituting around 4% of the dataset. This test subset served as a benchmark to evaluate the model's performance after training.

The implementation of the model was realized through the utilization of Ultralytics (Jocher et al. 2023), an open-source object detection and classification framework. The coding environment and computational resources used were the same as for the face detector. The training process was made with a batch size of 16 frames per batch and a 0.01 learning rate. The training persisted for 100 epochs, corresponding to the number of times the complete dataset was processed, effectively amounting to a total of 37 225 batches.

*Social Network Analysis*

In the perspective of evaluating our future pipeline capacity to output a reliable production of social network analysis based on automatically generated co-occurrence matrix, we performed a first social network analysis based on manually computed co-occurrences of individuals in collected field videos. We considered dyadic association indices as the probability of observing the focal individual with another individual in the same video, as a proxy of physical proximity between them. These indices were computed as 'simple ratio' (Hoppitt & Farine, 2018) of the number of co-occurrences of individuals A and B, divided by the total number of occurrences of appearance of both A and B on the dataset (**supplementary material**). In order to describe the overall social network profile, we computed the following social network measures, using the Animal Network Toolkit Softare package (ANTs, (Sosa et al., 2020) on Rstudio (R Core Team, 2021): (i) the network density, representing the ratio between existing links and all potential links in the network (Pasquaretta et al., 2014; Sosa et al., 2021); and (ii) the global efficiency, reflecting how fast information can spread through the network with the minimum number of connections (Pasquaretta et al., 2014). Furthermore, for each individual we computed sociality values such as the degree (i.e., the total number of individuals with which the subject co-occurred), the strength (i.e., the sum of all association indices of the individual) and the eigenvector centrality, reflecting how much an individual is integrated in the network (Pasquaretta et al., 2014). The package igraph (Csardi & Nepusz, 2006) was then used to represent graphically the group social network, spatialized following a GEM layout algorithm.

**Results**

*Face Detector*

After training, our Japanese macaques face detection CNN model displayed an average precision of 82.2% for an Intersection over Union (IoU) of 0.5. It means that 82.2% of predicted detection boxes by the AI are superposing at least 50% on manually annotated boxes. As bounding boxes can't match precisely with the shape of the macaques' face and considering variability of face angles and point of views, the fact that two bounding boxes do not superpose a lot more than 50% while still showing the same information is expected. This AP50 (e.i. Average Precision at IoU 0.50) value for the face detector reflects a clear capacity to place bounding boxes on Japanese macaques' faces (Figure 3). The training phenomenon can be observed with the false negative rate evolution through batches on Figure 4, starting from a 100% false negative rate at first batch to 13,65% rate reached at the 5699[th] batch.

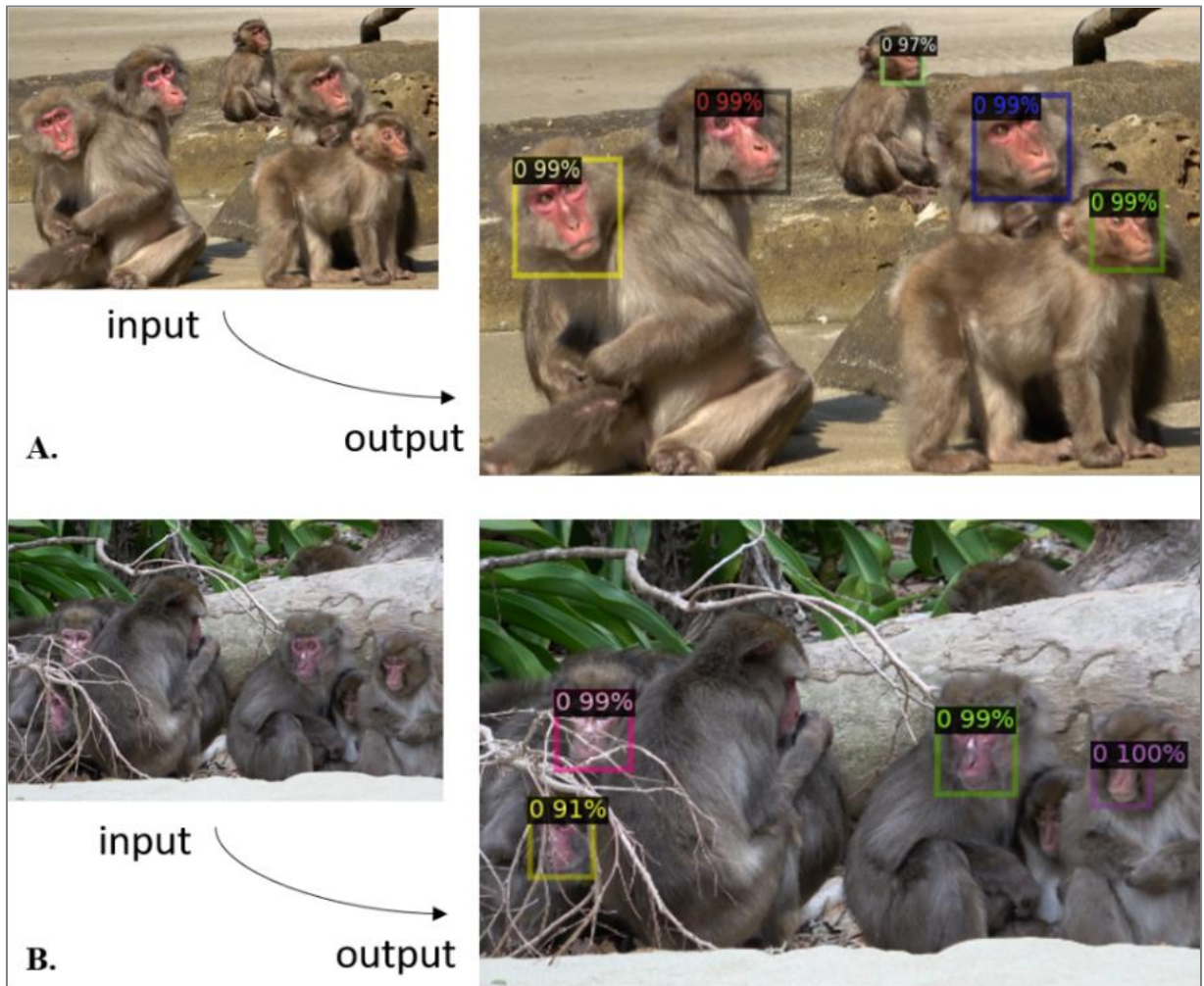

Figure 3. Examples of two frames before and after face detection by our AI model. We can see on frame (B) an example of false negative detection on a partially obstructed face.

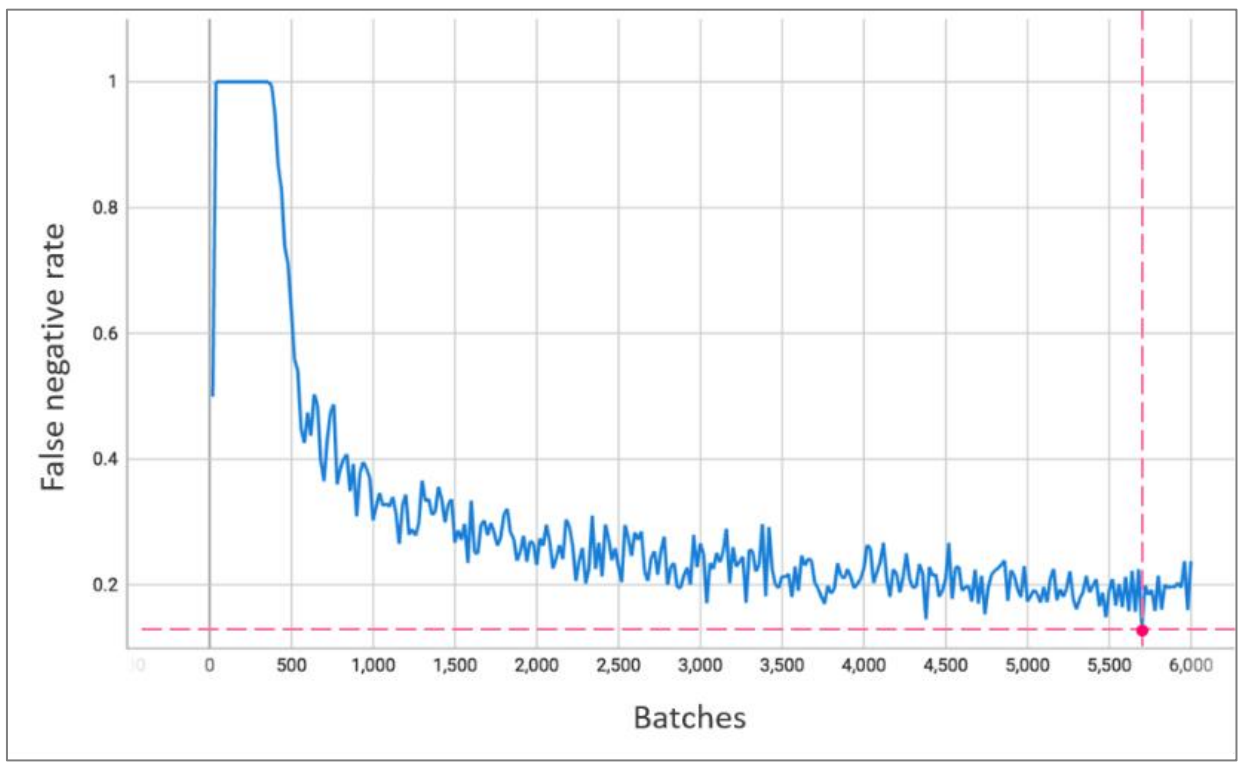

Figure 4. Curve of false negative rate of the face detection model depending on the number batches on which it was trained. A pink dot is placed at the lowest false negative rate reached (0.1365 at 5699[th] batch)

*Individual Classifier*

Our classifier model for individual recognition displayed a performance of 83% for its top1 accuracy and 92.92% for its top5 accuracy (Figure 5). When the AI predicts the identity, it proposes a probability score for every class (i.e., the 42 names). The top5 accuracy means that 92.92% of the time, our model predicts the good individual in its five most probable estimations. The 83% of top1 accuracy means the main prediction given by the AI is the good answer 83% of the time. This accuracy has been reached after 66 epochs. The normalized confusion matrix, expressing the proportion of predicted identity for every individual in the test set, underline a remarkably high rate of correct individual classification (Figure 6).

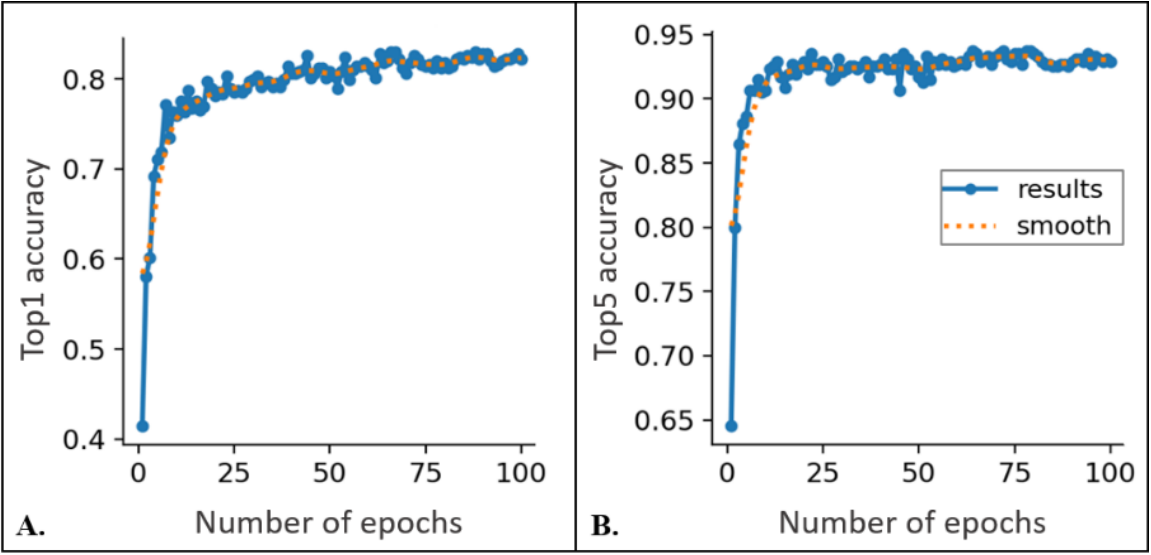

Figure 5. Curves of Top-1 (A) and Top-5(B) Accuracy of the individual classifier depending on the number epochs on which it trained.

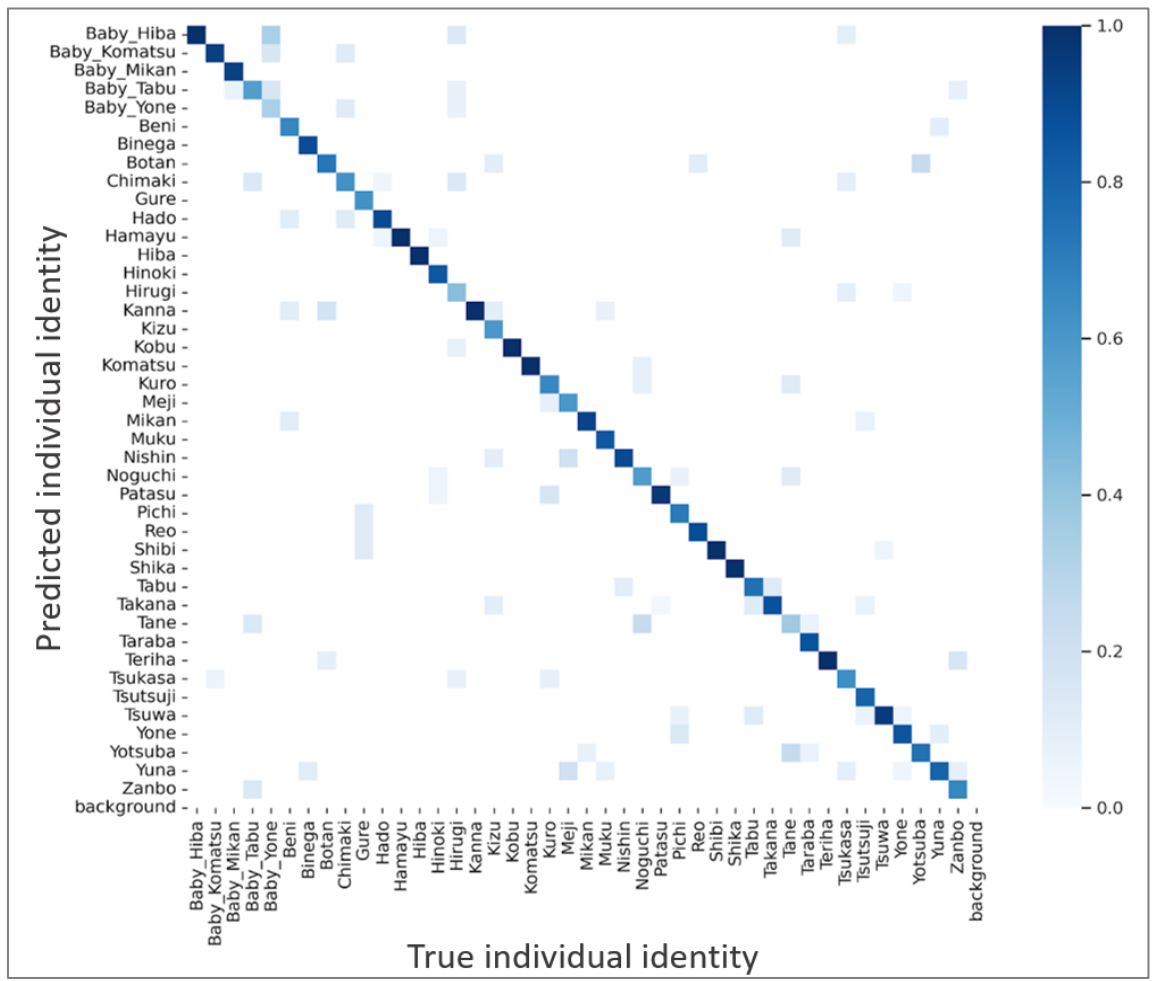

Figure 6. Normalized confusion matrix of predicted individual identity vs true individual identity.

A blue scale represents the percentage of predicted identities for each individual in the test set.

*Social Network Analysis*

A total of 276 dyadic co-occurrences of individuals on the videos have been observed. The resulting network of associations had a score of density of 0.173 and a global efficiency of 0.508. Network individual measures, i.e., the density, strength and eigenvector centrality of each subject are reported on Annexe 3. The individuals with the highest degree (D) and strength (S) were mostly juveniles such as Hirugi (D = 15, S = 1.180), Hinoki (D = 11, S = 1.011), Chimaki (D = 11, S = 1.200), Tsukasa (D = 11, S = 1.114) (Figure 7). Among the individuals with the highest eigenvector centrality indices (E), we can notice a remarkably high representation of mothers, such as Tsuwa (E = 0.895), Hiba (E = 0.885), Kizu (E = 0.884) and Yone (E = 0.822). Substantial association values were observed mainly between mothers and their offspring (e.g., 0.370 for Tabu and Baby_Tabu, 0.343 for Tsuwa and Kuro), even in adult dyads (e.g., 0.258 for Kanna and Yuna).

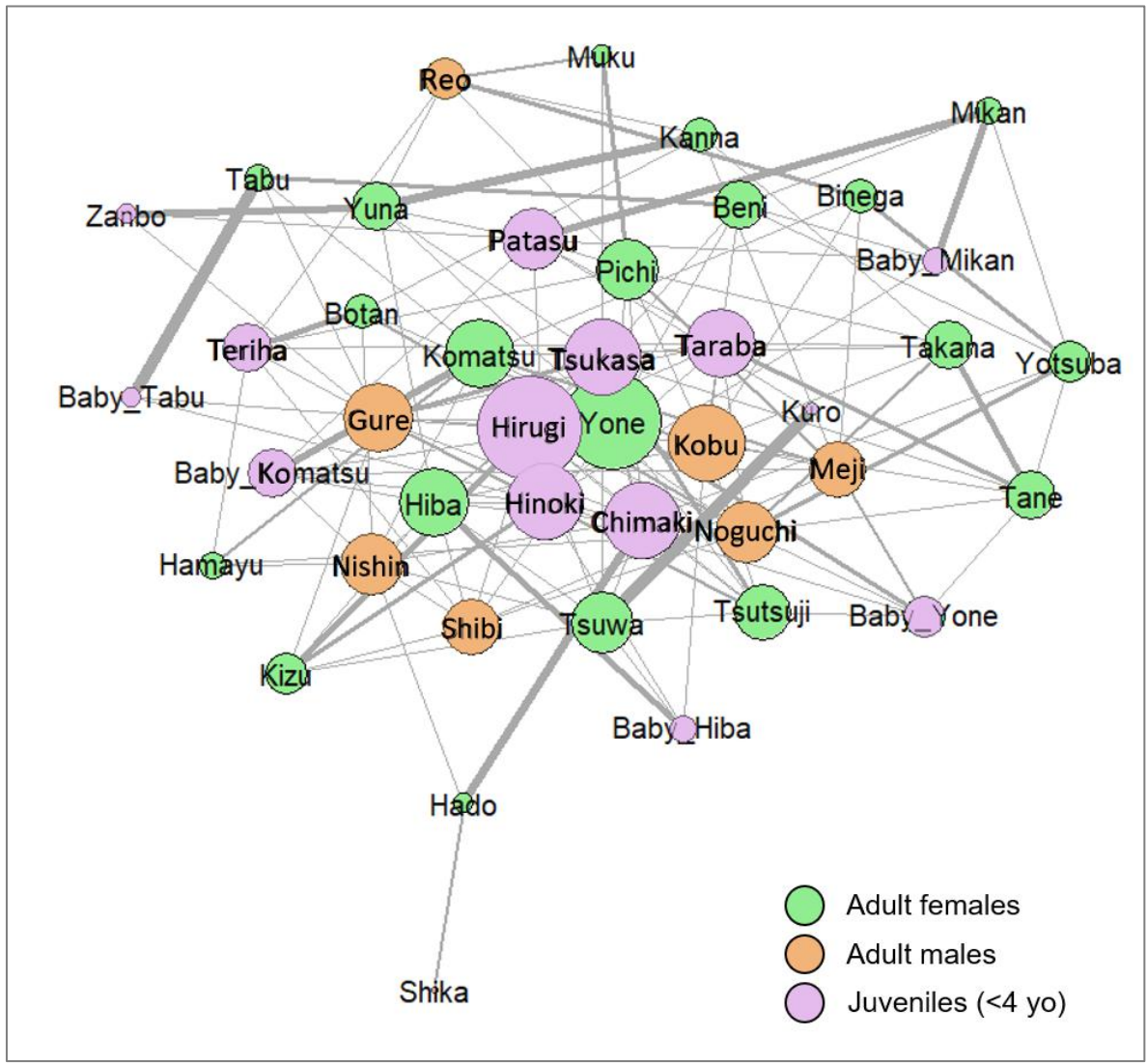

Figure 7. Network representation, following a GEM layout algorithm, of the co-occurrences of focal individuals with congeners on videos. Nod size represents the degree value for each individual and lines thickness between two nods represent the dyadic association index for the two individuals.

**Discussion**

Here we provide a first exploration to assess feasibility of the creation of a fully automatic pipeline for Japanese macaques' face detection and recognition from videos with deep learning followed by social network data generation. The primary focus of our work was the creation of a face detector for Japanese macaques and an identity classifier for Kōjima "main group" population, both unified in a fluid code pipeline, from video frames extraction to identity prediction (Figure 8).

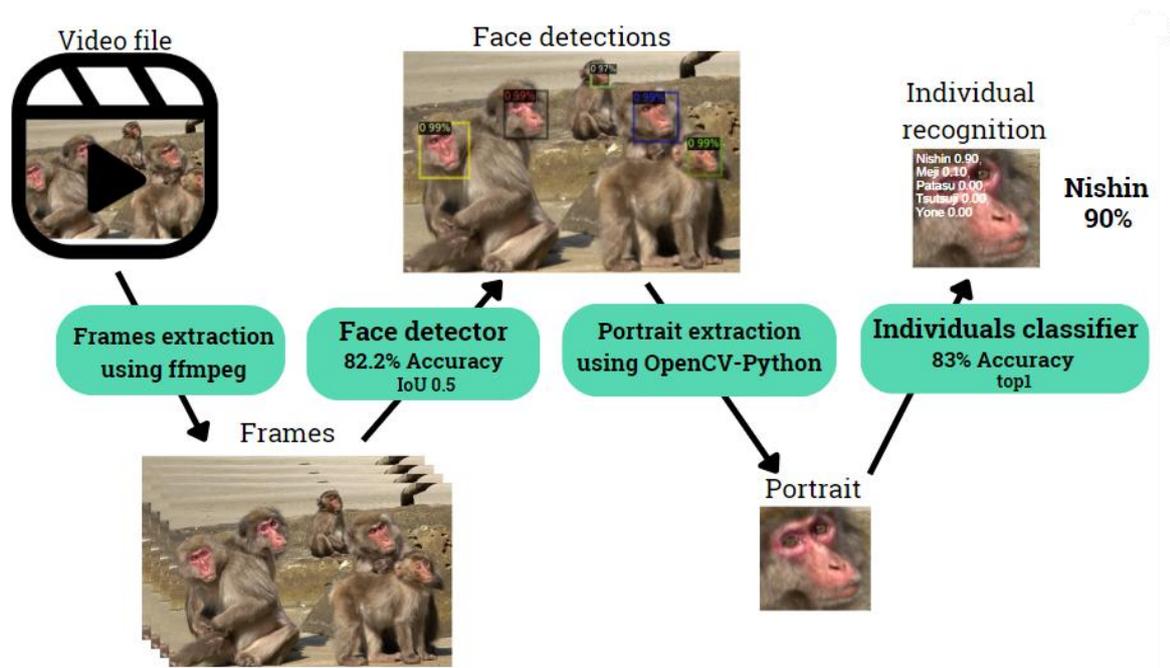

Figure 8. Chronological representation of the different codes of the pipeline applied on a frame and the different outputs generated.

Both the face detector (with an IoU threshold of 0.5) and the individual classifier achieved accuracy levels exceeding 80%, indicating a successful learning process for both models. This work represents the first operational deep learning-based AI pipeline designed for the study of Kōjima Island macaques. Consequently, we are currently working on an ergonomic intuitive Google Colab interface to make our code accessible to anyone. Utilizing this platform, researchers would be able to effortlessly incorporate their own data and execute program steps, subsequently visualizing outputs (i.e., face detection boxes on frames and identifications on portraits). Thus, beyond serving as a proof of concept for future models, this first tool could significantly benefit non-AI-expert researchers working with Kōjima Island macaques, facilitating various studies requiring individual recognition.

However, while this first step is a significant stride forward, there remains room for improvement. Indeed, our results are similar to some prior studies using deep learning and individual recognition (e.g. Schofield et al., 2019), others displayed stronger accuracy values with different approaches (Otani & Ogawa, 2022; Ueno et al., 2021). Several limitations surface within our method, each of which presents an opportunity for future research. Firstly, we must acknowledge that the composition of our datasets may have contributed to suboptimal results. Notably, we chose to annotate faces as long as they remained visible from a front view to a profile view. Faces are complex 3D objects and the coexistence of front and profile perspectives within the same category may have introduced ambiguity during the training process. It is plausible that excluding profile viewpoints from the dataset could lead to improved detection and recognition capabilities for front-facing images. However, this modification may also imply a decrease in the diversity of point of views the models can handle. Conducting further tests with different datasets could help us strike the optimal balance between accuracy and diversity in this context.

Technical limitations also bear consideration. Some hyperparameters were constrained due to resource limitations, e.g., a batch size capped at 10 frames per batch for the face detector training. As batch size can greatly impact training, this strongly limits the potential of exploration for best hyperparameters. Also, despite a comprehensive bibliography underpins our work, none of the research team members is an expert in artificial intelligence and deep learning. It is highly probable our naive approach on the subject over a short period of time made our current pipeline unoptimized. Consequently, considerable room remains for exploration and improvement of this dataset to yield optimal results. For this reason, we want to freely share our code through the upcoming Google Colab interface, making it easily accessible and editable. Thus, we invite any interested researchers with programming and deep learning skills to customize our code as needed and explore ways of improvement with the base we provide.

Considering we expect our code to produce automatic social network description based on our data, we manually made this social network description to prepare a performance assessment by comparison. We can observe that this network offers some reliable description of the social life of Japanese macaques as we know it. For example, we can see a remarkable inter-individual variability among adult females in their number of relationships, represented by the nods size. This is consistent with what we know of the species as there is a significant increase in social behaviors in females' Japanese macaques correlated with hierarchical position in the group (N. F. Koyama, 2003). Thus we could infer for example that Yone, who has the greatest nod among female adults, may occupy a high place in the hierarchy. Another good reliability point is the fact that the juveniles, apart from the youngest babies who always stay with their mother, are generally the most connected individuals in the troop. The only exception is Kuro, who is a young male about to turn 4 years old. Those dynamics are quite supported by Nakamichi studies from 1989 which depict juvenile females as very interactive with various individuals, while subadult males have a stronger tendency to go to the periphery of the troop. Anyway, those are only preliminary observations for a more in depth study of social networks with the future automatically generated network.

Nevertheless, we can identify flaws in our data collection method which can limit the relevance of our results. Overall, adverse weather conditions regularly hindered our ability to collect a substantial amount of data on the island. A number of 276 co-occurrences for 42 individuals is a good start to infer some social tendencies, but a larger amount of data would be needed to establish a more relevant and reliable co-occurrences dataset. Moreover, the two-month duration of data collection on Kōjima Island, spanning February and March, might not suffice to comprehensively elucidate the intricacies of the population's social network, knowing that Japanese macaques can exhibit variation in social behaviors depending on the period in the year. This fact can be exemplified by the absence of Shika, the troop alpha male, during most of the fieldwork period. As the fieldwork period overlapped with mating season, it's plausible that Shika sought out females from the other troop. We could reasonably think that Shika would have occupied a central position in our social network, interacting with several females, and exerting a significant impact on the social dynamics of the troop. Enhancing the representativity of our co-occurrence network could involve deploying data collection across different periods, thereby capturing the full complexity of social interactions exhibited along the year. While organizing fieldwork can be complex, one viable option to increase data quantity and period diversity could be to employ automatic data collection through the use of camera traps. However, transitioning to camera traps would involve changing the current co-occurrence measure method. Indeed, considering a co-occurrence on a frame as social interaction is simple but non-resilient to non-normed tapes. Wider angle views of camera traps, compared to our zoomed shots, would not allow us to use co-occurrence as a proxy for physical proximity. This is why we should explore other more complex interaction measurement methods. We could use an algorithm that considers the distance between detected bounding boxes and their size to express depth distance. These values could then be compared to a threshold to determine whether an interaction has occurred and should be recorded.

At its current stage, our AI pipeline is specialized for analyzing individual pictures one at a time. We are also working to make our pipeline capable of processing entire datasets of frames in a single run, thereby making it able to analyze large datasets. Thus, we could implement the automatic co-occurrence matrix generation at the end of the pipeline. For this upgrade, we plan to use a custom version of the Kanade-Lucas-Tomasi (KLT) face tracking code created by Schofield et al. in 2019 for their work on chimpanzees. With this add-on, our extracted frames will no longer be processed one by one through the face detector code. Instead, they would be analyzed simultaneously by a face tracker, using our face detector model's values. Consequently, the forthcoming integration of the tracker into our code will serve as the last part of our pipeline and will thus enable fully automated identity recognition from input videos. Furthermore, by establishing face tracks spanning multiple frames within a video, our identity classifier will be able to make predictions based on several frames, substantially enhancing the reliability of our identification predictions.

We expect our work to set the stage for long-term longitudinal studies with Kōjima Island macaques. Through the pipeline we are providing with this project, it becomes easily conceivable to conduct follow-up research on the social network of this population over the years. Additionally, on the user-friendly Google Colab page we will share, we are planning to include an option to easily fine-tune the model with new datasets. This means our current model could be updated and re-trained by anyone using newly annotated data from the Kōjima population. Such fine-tuning would not only enhance the precision of the individual classifier but also allow for recognition of newcomers in the troop, such as babies and free-ranging males, by learning new identities. What makes this even more interesting is that fine-tuning could also be extended to other Japanese macaque populations. Researchers would have to annotate identities on the bounding box generated by our face detector. This way, the new training dataset for the classifier would lead the produced model to encompass new populations to its repertory of known individuals. Thus, in the future we could expect the development of a unified standardized accessible tool tailored for monitoring the individuals from the most extensively studied Japanese macaque populations (e.g., Awajishima, Arashiyama, Jigokudani, Yokushima). Such a tool would open up exciting opportunities for comparative studies between different populations. The examination of how social dynamics manifest differently within the same species across various populations would be particularly exciting. Japanese macaques have frequently demonstrated a remarkable propensity for cultural diversity, as seen in behaviors like potato washing, bathing in hot springs or playing with rocks. As a result, the study of social networks holds pivotal importance, given its role in shaping the transmission of cultural knowledge throughout the group. Moreover, it's plausible to speculate that social structures could be influenced by environmental factors and population-specific cultural traits. Thus, undertaking comparative studies on social structure could yield valuable insights. Unveiling the diverse facets of social networks and identifying the potential drivers behind such diversity offers a particularly appealing perspective.

In conclusion, we have successfully developed the main components of our AI pipeline project, including a robust face detector and an identity classifier tailored for Kōjima 'main group' macaques. The remaining steps involve handling large datasets with a face tracker and implementing automatic co-occurrence matrix generation using classifier outputs, which we are currently working on. Subsequently, we will evaluate its capacity to efficiently generate reliable automatic social network descriptions. Our goal is ultimately to make our code available to the scientific community. Importantly, technologies like the one used in this study offer not only powerful research tools but also avenues for conservation, science popularization, and raising public awareness.

By packaging our recognizer into a user-friendly software, we can enable real-time applications using smartphone cameras, similar to what Witham did in 2017 with webcams. Given that several studied

Japanese macaque populations are popular tourist attractions, providing accessible facial recognition tools could serve both entertainment and educational purposes. This approach has the potential to enhance the general understanding of artificial intelligence while highlighting the concept of individuality in wild animals. Raising awareness about individuality in wild animals is not just an opportunity to share ethological knowledge; it's also a means to foster empathy and improve our understanding of animal welfare issues. In the broader context, our tool opens up exciting prospects for accessible individual recognition and has the potential for diverse expansions through deep learning. This would stimulate fundamental research with Japanese macaques and offer opportunities to explore their social behaviors and interactions in new and meaningful ways.

**References**

Csardi, G., & Nepusz, T. (2006). The igraph software package for complex network research. *InterJournal, Complex Systems*, *1695*. https://igraph.org

Hoppitt, W. J. E., & Farine, D. R. (2018). Association indices for quantifying social relationships : How to deal with missing observations of individuals or groups. *Animal Behaviour*, *136*, 227‑238. https://doi.org/10.1016/j.anbehav.2017.08.029

Iwamoto, T. (1974). A bioeconomic study on a provisioned troop of Japanese monkeys (Macaca fuscata fuscata) at koshima islet, Miyazaki. *Primates*, *15*(2), 241‑262. https://doi.org/10.1007/BF01742286

Koyama, N. F. (2003). Matrilineal Cohesion and Social Networks in Macaca fuscata. *International Journal of Primatology*.

Otani, Y., & Ogawa, H. (2021). Potency of Individual Identification of Japanese Macaques (Macaca fuscata) Using a Face Recognition System and a Limited Number of Learning Images. *Mammal Study*, *46*(1), 85‑93. https://doi.org/10.3106/ms2020-0071

Pasquaretta, C., Levé, M., Claidière, N., van de Waal, E., Whiten, A., MacIntosh, A. J. J., Pelé, M., Bergstrom, M. L., Borgeaud, C., Brosnan, S. F., Crofoot, M. C., Fedigan, L. M., Fichtel, C., Hopper, L. M., Mareno, M. C., Petit, O., Schnoell, A. V., di Sorrentino, E. P., Thierry, B., … Sueur, C. (2014). Social networks in primates : Smart and tolerant species have more efficient networks. *Scientific Reports*, *4*(1), Article 1. https://doi.org/10.1038/srep07600

Sosa, S., Puga-Gonzalez, I., Hu, F., Pansanel, J., Xie, X., & Sueur, C. (2020). A multilevel statistical toolkit to study animal social networks : The Animal Network Toolkit Software (ANTs) R package. *Scientific Reports*, *10*(1), Article 1. https://doi.org/10.1038/s41598-020-69265-8

Sosa, S., Sueur, C., & Puga-Gonzalez, I. (2021). Network measures in animal social network analysis : Their strengths, limits, interpretations and uses. *Methods in Ecology and Evolution*, *12*(1), 10‑ 21. https://doi.org/10.1111/2041-210X.13366

Ueno, M., Kabata, R., Hayashi, H., Terada, K., & Yamada, K. (2022). Automatic individual recognition of Japanese macaques (Macaca fuscata) from sequential images. *Ethology*, *128*(5), 461‑ 470. https://doi.org/10.1111/eth.13277

**Supplementary material**

## Annex 1: Individual characteristics

| Name | Sex | Age (yo) | Name | Sex | Age (yo) |
|------|-----|----------|------|-----|----------|
| Muku | female | 18 | Komatsu | female | 7 |
| Kanna | female | 18 | Yotsuba | female | 7 |
| Gure | male | 17 | Takana | female | 6 |
| Kizu | female | 16 | Hamayu | female | 5 |
| Kobu | male | 15 | Noguchi | male | 4 |
| Mikan | female | 15 | Tane | female | 4 |
| Shibi | male | 15 | Hinoki | female | 3 |
| Meji | male | 14 | Kuro | male | 3 |
| Nishin | male | 14 | Patasu | male | 3 |
| Shika | male | 14 | Teriha | female | 3 |
| Beni | female | 13 | Taraba | female | 2 |
| Reo | male | 13 | Tsukasa | female | 2 |
| Yone | female | 12 | Chimaki | female | 1 |
| Yuna | female | 12 | Hirugi | female | 1 |
| Tsuwa | female | 11 | Zanbo | female | 1 |
| Binega | female | 9 | Baby_Komatsu | ? | 0 |
| Hiba | female | 8 | Baby_Yone | ? | 0 |
| Pichi | female | 8 | Baby_Tabu | ? | 0 |
| Tabu | female | 8 | Baby_Hiba | ? | 0 |
| Botan | female | 7 | Baby_Mikan | ? | 0 |

**Annex 2: Matrix of association indices** (simple ratio) based on individuals' co-occurrences in focal videos

| | Komatsu | Tabu | Yotsuba | Mikan | Yone | Baby_Mikan | Reo | Meji | Nishin | Kanna | Teriha | Hado | Hirugi | Tsukasa | Tsuwa | Pichi | Gure | Tsutsuji | Taraba | Chimaki | Zanbo | Baby_Hiba | Kizu | Yuna | Baby_Komatsu | Hinoki | Noguchi | Shibi | Baby_Tabu | Baby_Yone | Kobu | Hiba | Takana | Binega | Beni | Hamayu | Patasu | Botan | Tane | Kuro | Muku | Shika |
|---|---|---|---|---|---|---|---|---|---|---|---|---|---|---|---|---|---|---|---|---|---|---|---|---|---|---|---|---|---|---|---|---|---|---|---|---|---|---|---|---|---|---|
| **Komatsu** | | 0.03 | | 0.03 | | | 0.09 | 0.03 | | | | | 0.02 | | | | | | | | | | | | 0.23 | | 0.03 | | | | | 0.03 | 0.04 | | 0.08 | | | | | | | |
| **Tabu** | 0.03 | | | | | | | | | | | | | | | | | | | 0.03 | | | | | | | | | 0.37 | | | | | | 0.13 | | | | | | | |
| **Yotsuba** | | | | 0.05 | | | | | | | | | | | | | | | | 0.03 | | | | | | | 0.13 | | | | | | | 0.16 | 0.05 | | | | | | 0.08 | |
| **Mikan** | | | 0.05 | | | 0.22 | | | | | | | | | | | | | | | | | | | | | | | | | | | | 0.05 | 0.20 | | | | | | | |
| **Yone** | 0.03 | | | | | | | 0.06 | | | | | 0.02 | | | 0.03 | 0.03 | 0.03 | | | | 0.03 | 0.03 | | | | | 0.04 | | 0.14 | 0.03 | | | | 0.04 | | | 0.04 | | | | |
| **Baby_Mikan** | | | | 0.22 | | | | | | | | | | | | | | | | | | | | | | 0.04 | | | | | | | | | 0.06 | 0.05 | | | | | | |
| **Reo** | | | | | | | | 0.05 | 0.06 | | | | | | | | 0.05 | | | | | | | 0.04 | | | | | | | | | | 0.13 | | | | | | | | 0.09 |
| **Meji** | 0.09 | | 0.06 | | | | | | 0.04 | | | | | | | 0.11 | | | | | | | | | | 0.07 | 0.04 | | | 0.08 | | 0.04 | | | | | | | | | | |
| **Nishin** | 0.03 | | | | | | | | | | | | | 0.05 | | 0.03 | 0.03 | | | 0.07 | | | | | | 0.04 | | 0.05 | | | | | | | 0.07 | | | | 0.06 | | | |
| **Kanna** | | | | | | | 0.05 | 0.04 | | | | | | | | | | | | | | | | 0.26 | | | | | | | | | | 0.04 | | | | | 0.05 | | | |
| **Teriha** | | | | | | | 0.06 | | | | | | | | | 0.05 | 0.04 | | 0.04 | | | | | | | | | 0.05 | | | | | | | 0.07 | 0.18 | | | | | | |
| **Hado** | | | | | | | 0.05 | | | | | | | | | | | | | 0.24 | | | | | | | | | | | | | | | | | | | | | | 0.10 |
| **Hirugi** | 0.02 | | 0.02 | | | | | | | | | | | 0.02 | 0.04 | 0.02 | 0.02 | 0.02 | | | | | 0.17 | | | 0.05 | | | 0.03 | 0.03 | 0.03 | 0.03 | 0.06 | | | | | | | | | |
| **Tsukasa** | | | | | | | | 0.03 | | | | | | | 0.02 | 0.02 | | 0.13 | 0.14 | 0.06 | 0.03 | | 0.03 | | | | | | | | | | | | | 0.03 | | | 0.03 | | 0.04 | |
| **Tsuwa** | | | | | | | | 0.03 | | | | | | | | 0.04 | 0.02 | | 0.05 | 0.02 | 0.03 | 0.02 | | | | | | | | | 0.02 | | | | | | | | | | 0.34 | |
| **Pichi** | | | | | | | | 0.05 | 0.11 | | 0.05 | | | | | | 0.03 | | | | | | | | | | | | | 0.04 | 0.04 | | | | 0.04 | | | | | | 0.13 | |
| **Gure** | | 0.03 | | | | | | | 0.03 | | | 0.04 | | 0.02 | 0.13 | 0.09 | | 0.06 | 0.08 | 0.04 | 0.03 | | | | | | | | | | | | | | | | | | | | | |
| **Tsutsuji** | | | 0.03 | | | | | | | | | | | 0.02 | 0.14 | 0.05 | 0.09 | | | 0.03 | | | | | | | | | | 0.04 | 0.07 | | | | | | | | | | | |
| **Taraba** | | | | | | | | | | | | | | 0.04 | 0.02 | 0.06 | 0.06 | | | | | | | | | 0.03 | 0.03 | | | 0.03 | 0.04 | | | | 0.07 | 0.13 | | | | | | |
| **Chimaki** | | 0.03 | 0.03 | | | | | | 0.07 | | | 0.24 | 0.02 | 0.03 | 0.02 | 0.03 | 0.08 | 0.03 | | | | | | | | 0.03 | | | | | | | | | | | | | | | | |
| **Zanbo** | | | | | | | | | | | | | | | | | | | 0.04 | | | | | | 0.22 | | | | | | | | | | 0.05 | | | | | | | |
| **Baby_Hiba** | | | | | | | | | | | | | | | | | | 0.03 | | | | | | | | 0.04 | | | | | 0.05 | 0.19 | | | | | | | | | | |
| **Kizu** | | | | | | | | | | | | | 0.17 | 0.02 | | | | 0.03 | | | | | | | | 0.14 | 0.07 | | | | 0.07 | | | | | | | | | | | |
| **Yuna** | | | | 0.03 | 0.04 | | | | | 0.26 | | | 0.03 | | | | | | | 0.22 | | | | | | | | | | | 0.02 | | | | | | | 0.03 | 0.03 | | | |
| **Baby_Komatsu** | 0.23 | | 0.03 | | | | 0.07 | 0.04 | | | | | | | | | | | | | | | | | | 0.03 | | | | | | 0.03 | | | | | | | | | | |
| **Hinoki** | | | | 0.04 | | | | | | | | | 0.05 | | | | | | | 0.03 | 0.03 | | 0.04 | 0.14 | 0.03 | | | | 0.03 | 0.04 | 0.05 | | | | 0.03 | | | | | | | |
| **Noguchi** | 0.03 | | 0.13 | | | | | | | | | | | | | | | | | 0.03 | | | | | | | | 0.04 | | | | | 0.03 | 0.08 | | 0.06 | | | | 0.10 | 0.04 | |
| **Shibi** | | | | | 0.04 | | | 0.04 | 0.05 | | 0.05 | | 0.03 | | | | | | 0.07 | | | | | | | | 0.04 | | | | 0.06 | | | | | | | | | | | |

| | Komatsu | Tabu | Yotsuba | Mikan | Yone | Baby_Mikan | Reo | Meji | Nishin | Kanna | Teriha | Hado | Hirugi | Tsukasa | Tsuwa | Pichi | Gure | Tsutsuji | Taraba | Chimaki | Zanbo | Baby_Hiba | Kizu | Yuma | Baby_Komatsu | Hinoki | Noguchi | Shibi | Baby_Tabu | Baby_Yone | Kobu | Hiba | Takana | Binega | Beni | Hamayu | Patasu | Botan | Tane | Kuro | Muku | Shika |
|---|---|---|---|---|---|---|---|---|---|---|---|---|---|---|---|---|---|---|---|---|---|---|---|---|---|---|---|---|---|---|---|---|---|---|---|---|---|---|---|---|---|---|
| **Baby_Tabu** | | 0.37 | | | | | | | | | | 0.03 | | | | | | | | | | | | | | 0.03 | | | | | | | | | | | | | | | | |
| **Baby_Yone** | | | | 0.14 | | 0.08 | | | | | | 0.03 | | | | | | 0.04 | | | | | | | | 0.04 | | | | | | | | | | | | | 0.04 | | | |
| **Kobu** | | | 0.03 | | | | | | | | | 0.03 | | | | 0.04 | | 0.07 | 0.03 | | 0.05 | | | | | | | | | | 0.06 | 0.04 | 0.05 | 0.05 | | | | | 0.04 | | | |
| **Hiba** | | | | | | | | | | | | | 0.06 | | 0.02 | | | | | | | 0.19 | 0.07 | 0.02 | 0.03 | 0.05 | 0.03 | 0.06 | | | 0.06 | | | | | | | | | | | |
| **Takana** | 0.03 | | | | | | | | | 0.04 | | | | | | 0.04 | | 0.04 | | | | | | | | 0.08 | | | | | | 0.04 | | | | | | | 0.19 | | | |
| **Binega** | | 0.04 | 0.16 | | | | 0.13 | 0.04 | | | | | | | | | | | | | | | | | | | | | | | 0.05 | | | | | | | | | | | |
| **Beni** | | 0.13 | 0.05 | 0.05 | 0.04 | 0.06 | | | | | | | 0.03 | | | | | | | | | | | | | | | | | | 0.05 | | | | | | | | | | | |
| **Hamayu** | 0.08 | | | | | | | | 0.07 | | 0.07 | | | | | | | | | | | | | | | 0.06 | | | | | | | | | | | | | | | | |
| **Patasu** | | | 0.20 | | 0.05 | | | | | | | | | | | 0.04 | | 0.07 | | 0.05 | | | | | 0.03 | 0.03 | 0.03 | | | | | | | | | | | | | | 0.04 | |
| **Botan** | | | | 0.04 | | | | | | 0.06 | 0.05 | 0.18 | | | | | | | | | | | | | | 0.10 | | | | | | | | | | | | | | | | |
| **Tane** | | | 0.08 | | | | | | | | | 0.03 | | | | | | 0.13 | | | | | | | | 0.04 | | | | 0.04 | 0.04 | | 0.19 | | | | | | | | | |
| **Kuro** | | | | | | | | | | | | | | 0.34 | | | | | | | | | | | | | | | | | | | | | | | | | | 0.04 | | |
| **Muku** | | | | | | | 0.09 | | | | | 0.04 | | | 0.13 | | | | | | | | | | | | | | | | | | | | | | | | | | | |
| **Shika** | | | | | | | | | | | | | | 0.10 | | | | | | | | | | | | | | | | | | | | | | | | | | | | |

## Annexe 3: Individual network measures

| Individual | Degree | Strength | Eigenvector centrality |
|---|---|---|---|
| Baby_Hiba | 4 | 0.619 | 0.542 |
| Baby_Komatsu | 7 | 0.917 | 0.760 |
| Baby_Tabu | 3 | 0.862 | 0.731 |
| Baby_Yone | 6 | 0.732 | 0.612 |
| Beni | 7 | 0.809 | 0.605 |
| Binega | 5 | 0.836 | 0.550 |
| Botan | 5 | 0.831 | 0.545 |
| Baby_Mikan | 4 | 0.748 | 0.527 |
| Chimaki | 11 | 1.200 | 0.810 |
| Gure | 10 | 1.050 | 0.860 |
| Hado | 3 | 0.788 | 0.484 |
| Hamayu | 4 | 0.547 | 0.386 |
| Hiba | 10 | 1.202 | 0.885 |
| Hinoki | 11 | 1.011 | 0.814 |
| Hirugi | 15 | 1.180 | 1.000 |
| Kanna | 5 | 0.879 | 0.645 |
| Kizu | 6 | 0.993 | 0.884 |
| Kobu | 11 | 0.945 | 0.712 |
| Komatsu | 10 | 1.203 | 0.860 |
| Kuro | 2 | 0.760 | 0.687 |
| Meji | 8 | 1.049 | 0.755 |
| Mikan | 4 | 1.035 | 0.671 |
| Muku | 3 | 0.515 | 0.334 |
| Nishin | 9 | 0.859 | 0.579 |
| Noguchi | 9 | 1.057 | 0.712 |
| Patasu | 9 | 1.065 | 0.746 |
| Pichi | 9 | 1.042 | 0.653 |
| Reo | 6 | 0.861 | 0.490 |
| Shibi | 8 | 0.758 | 0.600 |
| Shika | 1 | 0.200 | 0.100 |
| Tabu | 4 | 1.107 | 0.814 |
| Takana | 7 | 0.911 | 0.704 |
| Tane | 7 | 1.062 | 0.794 |
| Taraba | 10 | 1.014 | 0.813 |
| Teriha | 7 | 0.973 | 0.576 |
| Tsukasa | 11 | 1.114 | 0.874 |
| Tsutsuji | 8 | 0.931 | 0.792 |
| Tsuwa | 9 | 1.190 | 0.895 |
| Yone | 14 | 1.139 | 0.822 |
| Yotsuba | 6 | 0.975 | 0.665 |
| Yuna | 7 | 1.252 | 0.789 |
| Zanbo | 3 | 0.609 | 0.495 |